**Mapping Arctic Polynyas from Sea Ice Concentration Data Using Hierarchical Partitioning and Differentiated Region Growing: Sensitivity to Threshold and Product Choices**

Tianyu Zhang [1,2], Haiyi Ren [3*], Zhilun Zhang [4,5], Yang Hong [1,2], Chunlei Mi [6] , Xiao Cheng [4,5], Fengming Hui [4,5*]

[1] National Key Laboratory of Space Target Awareness,Beijing 101416,China.

[2] School of Information Science and Engineering, Space Engineering University, Beijing 101416, China.

[3] The State Key Laboratory of Space Information System and Integrated Application, Beijing Institute of Satellite Information Engineering, Beijing 100086, China.

[4] School of Geospatial Engineering and Science, Sun Yat-sen University, Southern Marine Science and Engineering Guangdong Laboratory (Zhuhai), Zhuhai 519082, China.

[5] Key Laboratory of Comprehensive Observation of Polar Environment (Sun Yat-sen University), Ministry of Education, Zhuhai 519082, China.

[6] Beijing Automotive Technology Center Co., Ltd., Beijing 101300, China

Corresponding author: Fengming Hui (huifm@mail.sysu.edu.cn) ; Haiyi Ren (haiyiren@mail.bnu.edu.cn )

**Abstract**

Accurately delineating Arctic coastal polynyas from sea ice concentration data remains challenging because of complex coastal geometry and spatially heterogeneous ice conditions. We develop a daily sea ice concentration (SIC) mapping framework based on hierarchical partitioning, connected-component analysis, and differentiated region growing for polynya and open-water regions. Applied to September-May ice seasons from 2002/03 to 2024/25, the framework achieves a mean overall accuracy of 96.74% and a mean kappa coefficient of 0.93 in the 63-scene visual assessment. Sensitivity experiments reveal a consistent pattern: the absolute magnitude of SIC-defined polynya extent is strongly configuration-dependent, whereas major temporal features are comparatively more stable. Relative to a 75% upper SIC threshold, seasonal extent integrals differ by 13.48%, 32.47%, and 65.04% at thresholds of 80%, 85%, and 90%, respectively, while the broad interannual structure remains recognizable. The response is also strongly regional and seasonal, with the largest threshold sensitivity in the Central Arctic Ocean, very weak sensitivity in the Gulf of St. Lawrence, and greater pan-Arctic curve separation from January onward. At an 80% threshold, product-comparison MAPE is 68.19% between the 6.25 and 25 km products and 11.47% between the 12.5 and 25 km products under their respective comparison periods and reference definitions. Regional product disagreement is likewise heterogeneous. Together, these results show that reproducible long-term polynya monitoring requires a fixed and explicitly documented combination of SIC product, threshold, and spatial classification rules, while broad temporal patterns can remain informative even when absolute extent differs substantially.

**Keywords:** Arctic polynya; sea ice concentration; connected components; region growing; threshold sensitivity; passive microwave remote sensing.

## 1 Introduction

Arctic polynyas expose open water or thin ice within the surrounding ice cover. During freezing conditions, repeated ice formation in these openings contributes to oceanic heat loss and brine release. Satellite estimates of active frazil and ice production demonstrate their importance across Northern Hemisphere coastal seas[1]. Understanding these processes requires observations of where openings occur, how their boundaries evolve, and how long they persist. Long satellite records now support analyses of Antarctic polynya trends and interdecadal variability[2]. Polynya detection also supports assessments of Southern Ocean and Arctic climate simulations[3], [4]. These applications make a reproducible spatial definition essential: changes in a mapped boundary must have a clear relationship to the detection procedure that generates it.

Passive microwave sea ice concentration (SIC) provides frequent, large-area observations for long-term mapping, including during polar darkness and cloudy conditions. However, coarse sampling mixes ice and water near narrow openings and their boundaries, as recent work north of Greenland demonstrates[5]. Comparisons against Landsat observations also reveal differences among SIC products across the concentration range[6]. A low SIC value alone cannot distinguish a polynya from the marginal ice zone (MIZ) or the open ocean. These features can occupy overlapping concentration ranges. Their separation therefore requires information about the surrounding ice, coastline, and connection to extensive open water[7]. Notably, a mapped SIC boundary identifies a low-concentration feature and does not directly measure ice thickness or production.

Recent improvements in spatial detail broaden the available mapping tools. Synthetic aperture radar (SAR) segmentation of Sentinel-1 imagery maps Arctic ice and water at scales of a few hundred meters[8]. In the western Ross Sea, SAR-based delineation resolves narrow polynyas that coarser passive microwave observations can miss[9]. Super-resolution processing also refines the representation of polynya boundaries and leads north of Greenland[5]. Nevertheless, fine sampling does not remove all classification ambiguity. Deep-learning detection of small landfast-ice polynyas near Canadian communities still struggles when water and surrounding ice have weak backscatter contrasts[10]. Thus, spatial detail and reliable feature discrimination remain related but distinct requirements.

Multisensor methods address some of these ambiguities by combining complementary observations. Fusion of SAR and passive microwave measurements supports pan-Arctic SIC retrieval[11]. Multitask networks can also estimate concentration, ice development stage, and floe size within one framework[12]. Their performance depends on more than network architecture. Controlled input comparisons show that microwave channels and ancillary time, location, and weather information affect sea-ice mapping accuracy[13]. These developments improve the observational basis for detection. They also reinforce the need to distinguish the quality of an input SIC field from the subsequent rules that classify a feature as a polynya.

Automatic polynya methods now explicitly address spatial context and temporal continuity. Region growing already separates ocean-connected water and the MIZ in Northern Hemisphere frazil mapping[1]. Daily object tracking and persistence criteria support the Daily Edge of Each Polynya in Antarctica boundary record[14]. For Arctic SIC imagery, a hybrid U-Net and Gaussian mixture framework identifies candidate polynyas and removes ocean and MIZ false positives[7]. A geomorphon approach instead interprets Antarctic SIC patterns as morphological depressions[15]. These methods demonstrate several viable ways to move beyond a concentration cutoff. Their different treatment of connectivity, shape, and persistence also makes the operational definition part of the resulting polynya record.

A central remaining issue is how much of the derived record is stable under plausible processing choices. Raising the upper SIC threshold necessarily enlarges the concentration range available to the classifier, but in a connectivity-based method it can also change whether neighboring components remain isolated or become connected. Threshold sensitivity is therefore not only an area-expansion problem; it can alter object topology and do so differently among regions and seasons. Recent polynya studies likewise report strong threshold dependence and seasonally varying divergence among detected extents[15], [16]. This motivates testing both the absolute magnitude of mapped extent and the persistence of its temporal structure across threshold settings, and whether this sensitivity varies among regions and seasons.

Input-product sensitivity poses a complementary question. Passive-microwave SIC products differ in retrieval construction, weather filtering, spatial sampling, and their response at intermediate concentrations[6], [17], [18]. Applying the same classifier to several products therefore tests whether a mapping conclusion depends on the observational input as well as on the threshold. Recent Antarctic studies show that threshold, grid, season, and product choice can substantially affect detected polynya statistics[16]. Kern et al.[19] further showed systematic differences between PSSM-derived polynya classes and higher-resolution AMSR-E 89-GHz SIC estimates in the Ross Sea and off the Adélie Coast. Examining threshold sensitivity within one product and product sensitivity at a fixed threshold provides two distinct views of the robustness of the resulting Arctic record.

Against this background, the present study links an explicit daily mapping procedure with a systematic sensitivity analysis across Arctic regions and ice seasons. We ask three questions. First, can hierarchical partitioning and differentiated region growing separate enclosed low-SIC openings from ocean-connected MIZ and open water with high classification agreement? Second, how strongly does the mapped extent change when the upper SIC threshold is varied while the input product is fixed? Third, how much does the result change when the SIC product is changed at a fixed threshold? The emphasis is on what remains stable and what changes across configurations, so that the resulting record can be interpreted and reproduced consistently.

The workflow first partitions SIC into high- and low-concentration domains and labels connected components. It then identifies open-water and candidate polynya seeds within the low-concentration domain. Differentiated region growing expands these seeds under separate neighborhood rules, while geographic exclusions address known coastal and inland-water artifacts. This hierarchy uses concentration to identify candidate pixels and spatial context to determine their final class. A fixed 15% seed threshold initializes the regions; a separate upper threshold controls the concentration range available to the extraction procedure.

We apply the framework to September–May seasons from 2002/03 to 2024/25 using the 6.25 km ARTIST Sea Ice (ASI) record. The unavailable 2011/12 season is excluded. Visual assessment covers 63 scenes and the polynya and MIZ classes. The threshold experiment compares upper limits of 75%, 80%, 85%, and 90% within the same SIC product. A second experiment applies an 80% upper threshold to three products: 6.25 km ASI, 12.5 km enhanced NASA Team (NT2), and the 25 km climate data record. Comparisons use the respective periods of data availability. Regional and interannual comparisons then examine whether similar temporal behavior accompanies differences in absolute extent. This analysis provides a practical basis for selecting and reporting configurations in subsequent Arctic polynya studies.

## 2 Data and study area

### 2.1 Sea ice concentration products

Daily SIC fields are taken from the four product entries in Table 1. The 6.25 km ASI series combines AMSR-E and AMSR2 observations[20], [21]. It provides the primary input for method development, threshold comparisons and product comparison. The 12.5 km input is the NT2 SIC field from AU_SI12 Version 1[22]. The 25 km series uses the NOAA/NSIDC climate data record G02202 Version 5[23]. The near-real-time record G10016 Version 4 extends this series through January–May 2025[24]. These product versions are used throughout the analysis.

**Table 1** SIC products used in the analysis.

<table>
<tr><th>Product and version</th><th>Grid spacing</th><th>Retrieval</th><th>Period used</th><th>Role</th></tr>
<tr><td>Bremen n6250 ASI v5.4[20], [21]</td><td>6.25 km</td><td>ASI</td><td>Sep. 2002–May 2025; 2011/12 omitted</td><td>Primary mapping, threshold tests and product comparison</td></tr>
<tr><td>AU_SI12 v1[22]</td><td>12.5 km</td><td>NT2</td><td>Sep. 2012–May 2025</td><td>Product comparison</td></tr>
<tr><td>G02202 v5[23]</td><td rowspan="2">25 km</td><td>CDR combining NT and Bootstrap</td><td>Sep. 2002–Dec. 2024</td><td rowspan="2">Product comparison</td></tr>
<tr><td>G10016 v4[24]</td><td>Near-real-time CDR</td><td>Jan.–May 2025</td></tr>
</table>

*Notes: Only September–May observations are analyzed. Dates identify the subsets used, rather than the complete catalog availability of each product. CDR: climate data record.*

ASI is based on high-frequency microwave polarization information and includes weather filtering[21]. The NOAA/NSIDC climate data record combines NASA Team and Bootstrap retrievals[23], [24], whereas AU_SI12 provides enhanced NASA Team, or NT2, SIC[22]. Thus, the input products differ in retrieval construction as well as nominal grid spacing. Here, 6.25, 12.5, and 25 km refer to grid spacing; they should not be interpreted as identical measures of the effective radiometric footprint.

**2.2 Spatial domain and ice seasons**

The analysis uses Northern Hemisphere polar stereographic grids with a latitude of true scale at 70°N[25]. Regional statistics use the NSIDC mask distributed by the Centre for Polar Observation and Modelling[26], [27] and shown in Figure 1. The analysis considers an aggregate pan-Arctic series together with fourteen regional masks: Sea of Okhotsk, Bering Sea, Hudson Bay, Gulf of St. Lawrence, Baffin Bay, Greenland Sea, Barents Sea, Kara Sea, Laptev Sea, East Siberian Sea, Chukchi Sea, Beaufort Sea, Canadian Arctic Archipelago, and Central Arctic Ocean. Notably, as shown in Figure1, the pan-Arctic domain includes the fourteen named regional masks and additional ocean-grid areas in pale blue. Consequently, the pan-Arctic total is not restricted to the sum of the fourteen regional totals.

An ice season runs from September to May of the following year. The study spans 2002/03–2024/25 ice seasons, with the 2011/12 season unavailable in the primary ASI record. Short gaps within otherwise available records are treated as described in Section 3.1. The 2011/12 gap is not reconstructed as a daily ASI observation series. The 12.5 km comparison begins in 2012/13, following the AMSR2 record selected for this analysis.

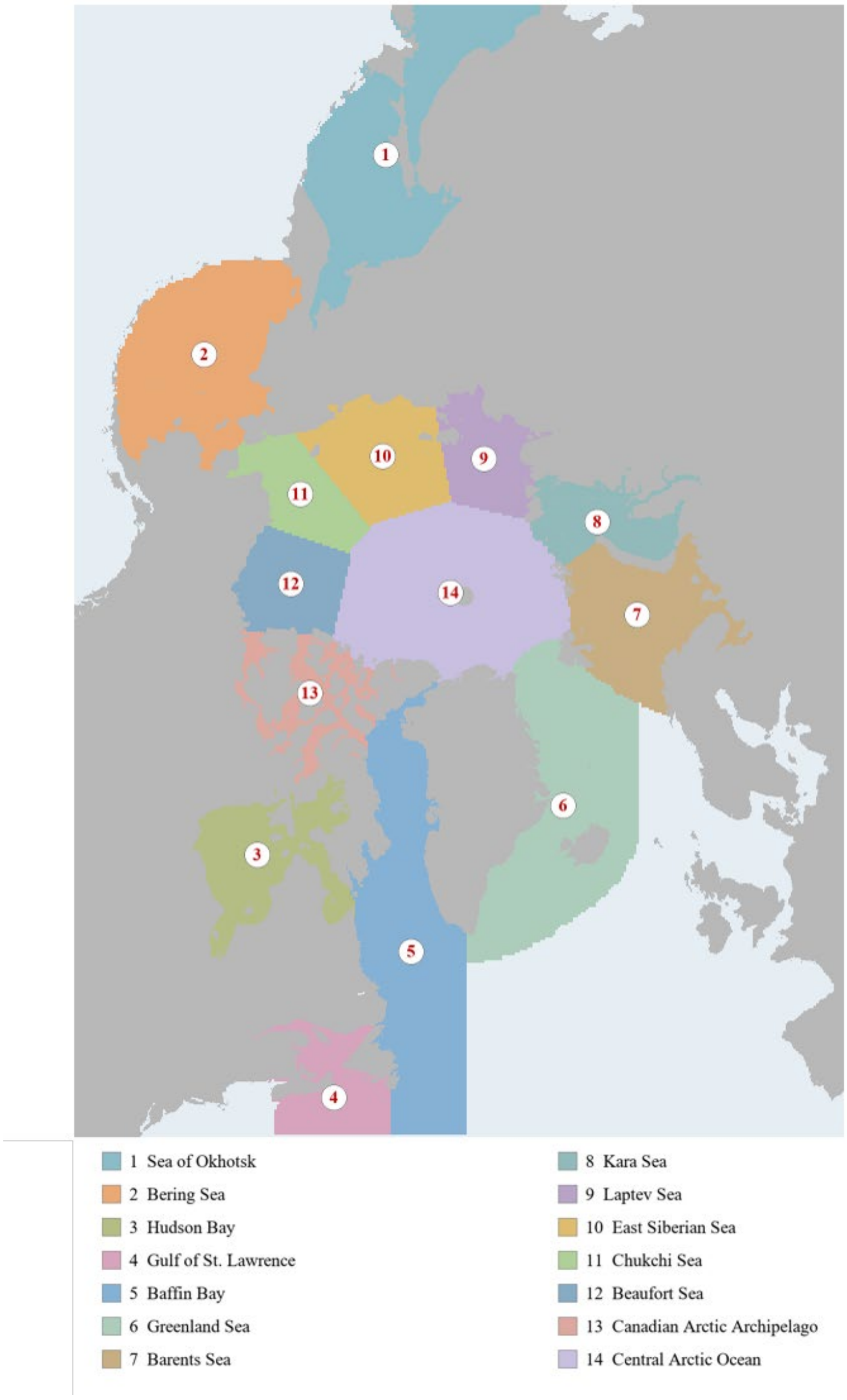

**Figure 1.** Regional mask used for the 6.25 km analysis, displayed from the existing categorical mask. Numbers identify the fourteen regional labels listed in the legend; gray denotes land or excluded cells and pale blue denotes other ocean-grid areas. The regional-mask source is [26].

## 3 Polynya extent extraction method

The mapping target is an SIC-defined low-concentration region that is separated from extensive open-ocean water by higher-concentration ice or land. The MIZ is the ocean-connected, nonzero-SIC portion of the low-concentration domain. This operational distinction depends on spatial connectivity, rather than SIC alone. Pixels above the upper threshold are referred to as high-concentration ice, since concentration does not determine ice thickness. Figure 2 summarizes the workflow, and Figure 3 illustrates the partition and classification stages.

**3.1 Data preprocessing**

Missing and extensively corrupted SIC scenes are first identified by visual inspection. Short gaps are filled using adjacent valid fields. For a missing date nearest to one end of a gap, the interpolation used in the workflow is expressed as:

$$C_{fill} = \frac{N}{N+1} C_{near} + \frac{1}{N+1} C_{far} \tag{1}$$

where N is the number of consecutive missing or invalid days, $C_{near}$ is the nearer valid SIC field, and $C_{far}$ is the farther valid field. For gaps spanning three or more days, reconstruction proceeds from the available outer dates toward the gap center. This procedure supplies a continuous input for short interruptions, but does not add independent information about opening or closing events during the missing dates.

Small patches of invalid coastal data were removed using manually delineated masks in the North Pacific and North Atlantic. These masks cover the Gulf of Alaska, southern Bering Sea, southern Sea of Okhotsk, and waters from eastern Newfoundland to the southern United Kingdom. Non-land pixels within these zones were assigned to seawater to prevent spurious polynya detection. The Bohai Sea was also masked and assigned to seawater during preprocessing.

Inland-water masks were constructed using the land surroundings of candidate water bodies. Neighboring land pixels were checked along their boundaries to identify enclosed lakes and rivers. Estuaries and rivers connected to the ocean were delineated manually. The resulting mask excluded these inland waters from polynya extraction.

These geographic exclusions are held fixed throughout the experiments. They remove known coastal and inland-water artifacts before seed identification and ensure that threshold comparisons are made over a consistent operational domain. Their implementation is transferred to each product grid in the product-sensitivity experiment.

**3.2 Hierarchical partitioning**

The first partition divides valid ocean pixels into high-concentration ice, with SIC

greater than the upper threshold TH, and a low-concentration domain with $0 \leq SIC \leq TH$. The reference configuration uses TH = 80%; the threshold experiment changes only this upper value to 75%, 80%, 85%, or 90%. Four-neighbor connected-component labeling is applied to the low-concentration domain. Up to the two largest connected components form the candidate ocean-connected domain, accommodating the major Atlantic and Pacific open-water regions when they are separate. The remaining components are retained as isolated low-concentration candidates.

Within the candidate ocean-connected domain, a fixed 15% threshold separates very-low-concentration pixels from the intermediate band, $15\% < SIC \leq TH$. Four-neighbor labeling is then applied to the very-low-concentration pixels. The largest components provide open-water seeds, while smaller detached components provide candidate polynya seeds. This lower threshold initializes the seed regions. It does not define the upper SIC limit of the final polynya mask and is held fixed in the sensitivity experiment.

Geographic exceptions prevent large, seasonally enclosed water bodies from being classified as polynyas solely because they are disconnected from the two largest components. The exceptions cover specified parts of the Baltic Sea, Sea of Okhotsk, northern Russian waters, and Hudson Bay. Components in these areas are assigned to open water when they exceed the reported area criterion of 150,000 $km^2$. This criterion excludes large water bodies; it does not impose a minimum polynya area.

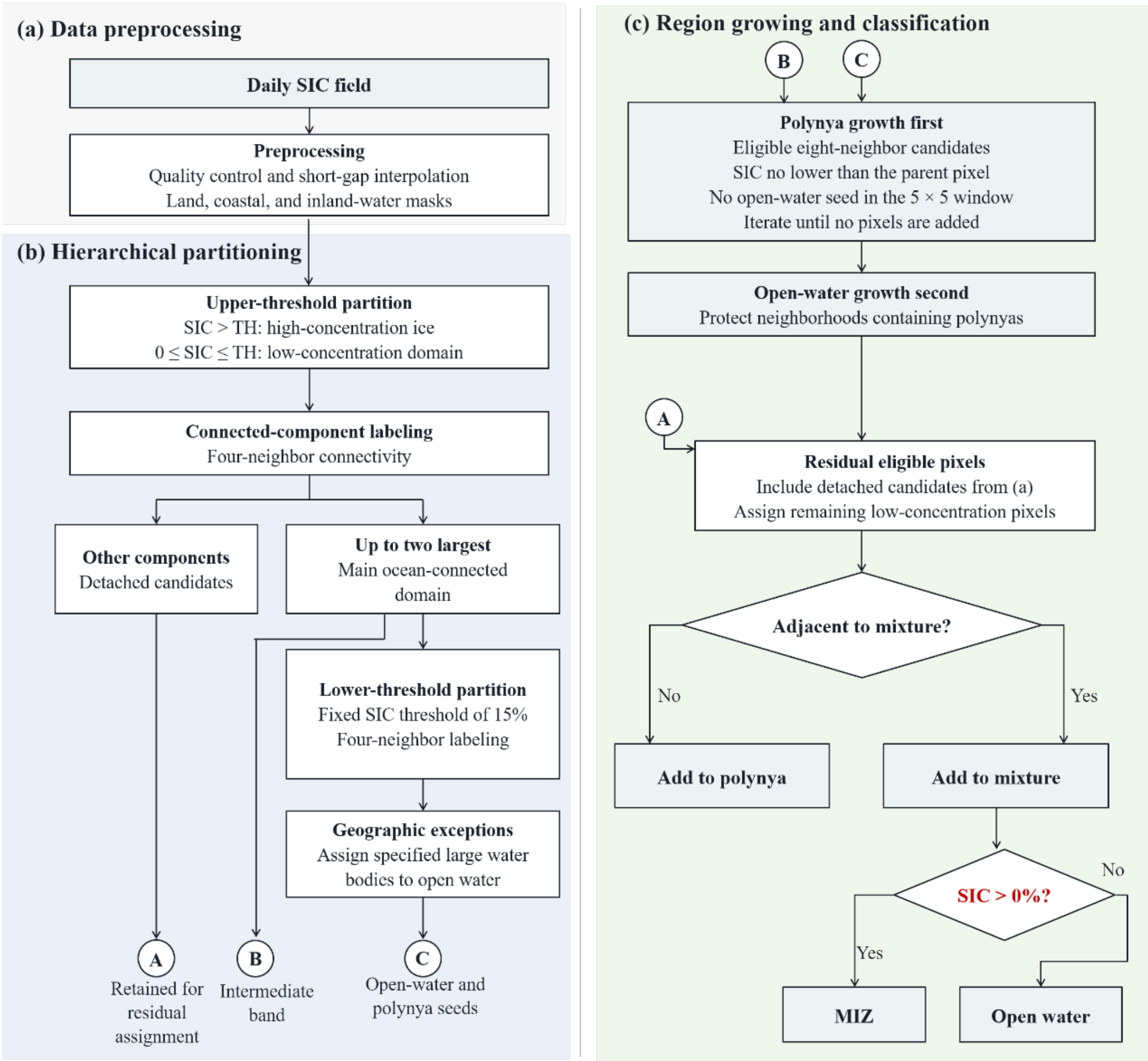

**Figure 2.** Workflow for Arctic polynya mapping: (a) Data preprocessing; (b) hierarchical SIC partitioning and seed identification; (c) sequential region growing and final classification. Connectors A, B, and C denote isolated low-concentration candidates, the intermediate-concentration band and seed regions, respectively. The reference upper SIC threshold is 80%, and the fixed seed threshold is 15%. The final SIC > 0% test applies only to ocean-connected mixture pixels.

### 3.3 Region growing and final classification

Boundary tracing identifies the initial growth pixels of the polynya and open-water seeds. Candidate polynya boundary pixels directly adjacent to open-water seeds are excluded from the initial polynya growth set. Polynya growth then evaluates neighboring pixels iteratively. An admissible candidate is neither land nor high-concentration ice, has not already been assigned to an excluded class, and has SIC no lower than that of its parent growth pixel. Its surrounding 5 × 5 neighborhood must contain no open-water seed pixel. New accepted pixels provide the boundary for the next iteration, and growth stops when no additional pixels are accepted.

Growth evaluates the eight cells surrounding a pixel in a 3 × 3 window. A wider separation check evaluates the twenty-four surrounding cells in a 5 × 5 window. The wider check limits expansion toward open-water seeds, while the eight-neighbor rule permits diagonal growth. Together, the two neighborhood scales allow local boundary growth while preserving separation from the larger ocean-connected domain.

Open-water growth follows polynya growth. It expands into eligible low-concentration neighbors while excluding land, high-concentration ice, and pixels already assigned to polynyas. Neighborhoods containing classified polynya pixels are protected from this expansion. The resulting ocean-connected region is provisionally an ice–water mixture. Remaining eligible pixels adjacent to that region are assigned to the mixture; the other residual low-concentration pixels, including isolated candidates from the first partition, are assigned to polynyas. Finally, mixture pixels with SIC > 0 are labeled MIZ, and those with SIC = 0 are labeled open water. Zero-SIC pixels within a polynya retain their polynya label.

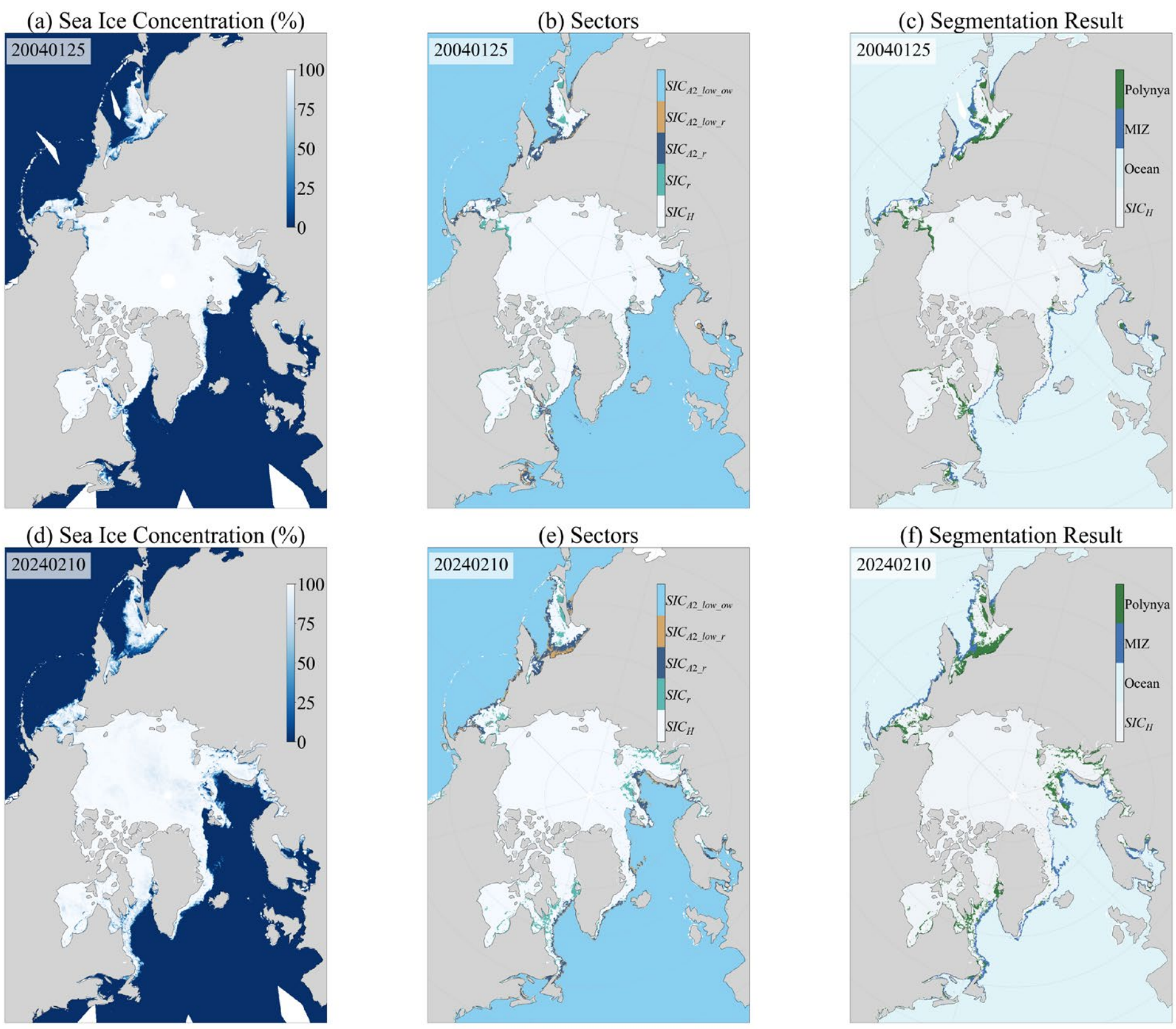

**Figure 3.** Examples of hierarchical partitioning and final classification on January 25, 2004 (a–c), and February 10, 2024 (d–f). Panels (a, d) show SIC, (b, e) show intermediate partitions, and (c, f) show the final classes. In the partition legends, $SIC_H$ denotes high-concentration ice, $SIC_r$ denotes the detached candidates, $SIC_{A2_r}$ denotes the intermediate-concentration band, $SIC_{A2_low_ow}$ and $SIC_{A2_low_r}$ denote the open-water and polynya seed regions, respectively.

### 3.4 Extent and comparison metrics

Daily polynya extent is the sum of the nominal areas of pixels classified as polynya, including both their open-water and ice-covered fractions:

$$\mathrm{E(t)} = \alpha N_t \tag{2}$$

where $N_t$ is the number of polynya pixels on day $\mathrm{t}$ and $\alpha$ is nominal pixel area, calculated as the square of grid spacing in the existing analysis. This is a classified extent, not an open-water area weighted by 1 − SIC. Summing daily extents over an ice season gives a seasonal extent integral:

$$I_s = \sum_{t=1}^{D} E(t)\Delta t \tag{3}$$

Here $\mathrm{D}$ counts analyzed days and $\Delta \mathrm{t} = 1$ day, giving $\mathrm{I}$ in km²·day. An integral weights both the size and persistence of mapped openings; it is not the geographical area of a unique footprint. Multi-season daily averages are used to describe how differences evolve through September–May. Month labels on these plots indicate position within the season and do not imply that each curve consists of nine monthly means.

For paired seasonal integrals X and Y, we calculated the mean absolute percentage error (MAPE):

$$MAPE = \frac{100\%}{K}\sum_{k=1}^{K}\left|\frac{Y_k - X_k}{X_k}\right| \tag{4}$$

$\mathrm{X}$ is the stated reference series, $\mathrm{Y}$ is the comparison series, and $\mathrm{K}$ is the number of valid paired seasons. MAPE is used as a difference measure between two products, without treating either product as ground truth. The threshold calculations use the lower-threshold series as the reference. For the product comparisons, the 6.25–25 km value uses 25 km as reference, whereas the 12.5–25 km value uses 12.5 km as reference. Different denominators and periods are retained explicitly rather than treated as interchangeable comparisons.

### 3.5 Visual assessment of the classification

The assessment uses stratified random sampling within the mapped polynya and MIZ classes, followed by visual interpretation. Stratification permits different sampling effort among classes, but the reference labels, inclusion probabilities, and estimation procedure must be specified for a population-level accuracy statement[28], [29]. The reported sampling setup assumes expected user's accuracies of 0.90 for polynyas and 0.85 for the MIZ, with a sampling-control parameter V = 0.01.

The visual assessment concerns the 6.25 km record reported at the 80% threshold. Seven scenes were selected from each September–May calendar month, giving 63 scenes from a stated total of 6,010. Table 2 presents monthly confusion counts and user's accuracy (UA), producer's accuracy (PA), overall accuracy (OA), and kappa. The mean OA is 96.74%, and mean kappa is 0.93. Mean UA/PA is 97.32%/95.94% for polynyas and 96.08%/97.49% for the MIZ.

The monthly confusion matrices reveal a consistent asymmetry that is more informative than the mean accuracy alone. Polynya producer's accuracy decreases to 92.22% in December and 93.15% in April, while the corresponding user's accuracies remain 99.07% and 97.44%. Under the table's class convention, these months therefore contain more omission than commission of polynya pixels, indicating that difficult cases tend to be absorbed into the competing MIZ class rather than generated as false polynyas. This pattern is consistent with the known difficulty of passive-microwave SIC retrieval and class separation at intermediate concentrations and mixed pixels[6],[17]. The high overall scores nevertheless show that the connectivity-based workflow maintains strong class agreement across the sampled scenes.

**Table 2.** Confusion Matrix for Polynya Extent Extraction.

| Month | Categories | Pixel Count | | User's Accuracy (%) | Producer's Accuracy (%) | Overall Accuracy (%) | Kappa Coefficient |
|---|---|---|---|---|---|---|---|
| | | Polynya | MIZ | | | | |
| Sep. | Polynya | 279 | 6 | 98.24 | 97.89 | 98.57 | 0.97 |
| | MIZ | 5 | 477 | 98.76 | 98.96 | | |
| Oct. | Polynya | 272 | 8 | 98.55 | 97.14 | 98.45 | 0.97 |
| | MIZ | 4 | 491 | 98.40 | 99.19 | | |
| Nov. | Polynya | 350 | 9 | 95.63 | 97.49 | 96.76 | 0.93 |
| | MIZ | 16 | 396 | 97.78 | 96.12 | | |
| Dec. | Polynya | 320 | 27 | 99.07 | 92.22 | 96.05 | 0.92 |
| | MIZ | 3 | 409 | 93.81 | 99.27 | | |
| Jan. | Polynya | 356 | 17 | 96.22 | 95.44 | 95.97 | 0.92 |
| | MIZ | 14 | 382 | 95.74 | 96.46 | | |
| Feb. | Polynya | 377 | 12 | 94.49 | 96.92 | 95.57 | 0.91 |
| | MIZ | 22 | 356 | 96.74 | 94.18 | | |
| Mar. | Polynya | 356 | 13 | 96.74 | 96.48 | 96.68 | 0.93 |
| | MIZ | 12 | 372 | 96.62 | 96.88 | | |
| Apr. | Polynya | 381 | 28 | 97.44 | 93.15 | 94.96 | 0.90 |
| | MIZ | 10 | 335 | 92.29 | 97.10 | | |
| May | Polynya | 439 | 15 | 99.55 | 96.70 | 97.64 | 0.95 |
| | MIZ | 2 | 264 | 94.62 | 99.25 | | |
| **Avg.** | Polynya | - | - | **97.32** | **95.94** | **96.74** | **0.93** |
| | MIZ | | | **96.08** | **97.49** | | |

## 4 Sensitivity analysis

### 4.1 Sensitivity to the upper SIC threshold

The threshold experiment applies TH = 75%, 80%, 85%, and 90% to the same 6.25 km ASI input for the 2002/03–2024/25 ice seasons. Figure 4 compares seasonal integrals for the aggregate series and the fourteen regional masks. The experiment therefore isolates the effect of changing the upper SIC threshold within one observational product.

Higher thresholds produce progressively larger seasonal integrals. Relative to 75%, the aggregate differences are 13.48%, 32.47%, and 65.04% at 80%, 85%, and 90%, respectively, and the reported mean absolute differences span approximately 2.49–12.02 × $10^7$ $km^2$·day. Across the tested range, the reported mean extent response is about 4.30% per one-percentage-point increase in threshold. More importantly, the widening separation among curves is not captured by one universal multiplicative correction: the effect of admitting higher-SIC pixels accumulates differently among seasons and regions.

Despite the large magnitude differences, the broad interannual structure in Figure 4 remains recognizable across all four thresholds: many relative highs, lows, and multi-year changes occur in similar portions of the record. The key result is therefore a separation between magnitude sensitivity and temporal-pattern stability. Threshold selection strongly affects how much polynya extent is mapped, but it does not erase the principal temporal organization of the record. Similar threshold studies have reported large changes in absolute polynya area while retaining common stages of seasonal development[15], [16]. This relative stability is valuable for long-term comparison, provided that absolute values from different threshold configurations are not treated as interchangeable.

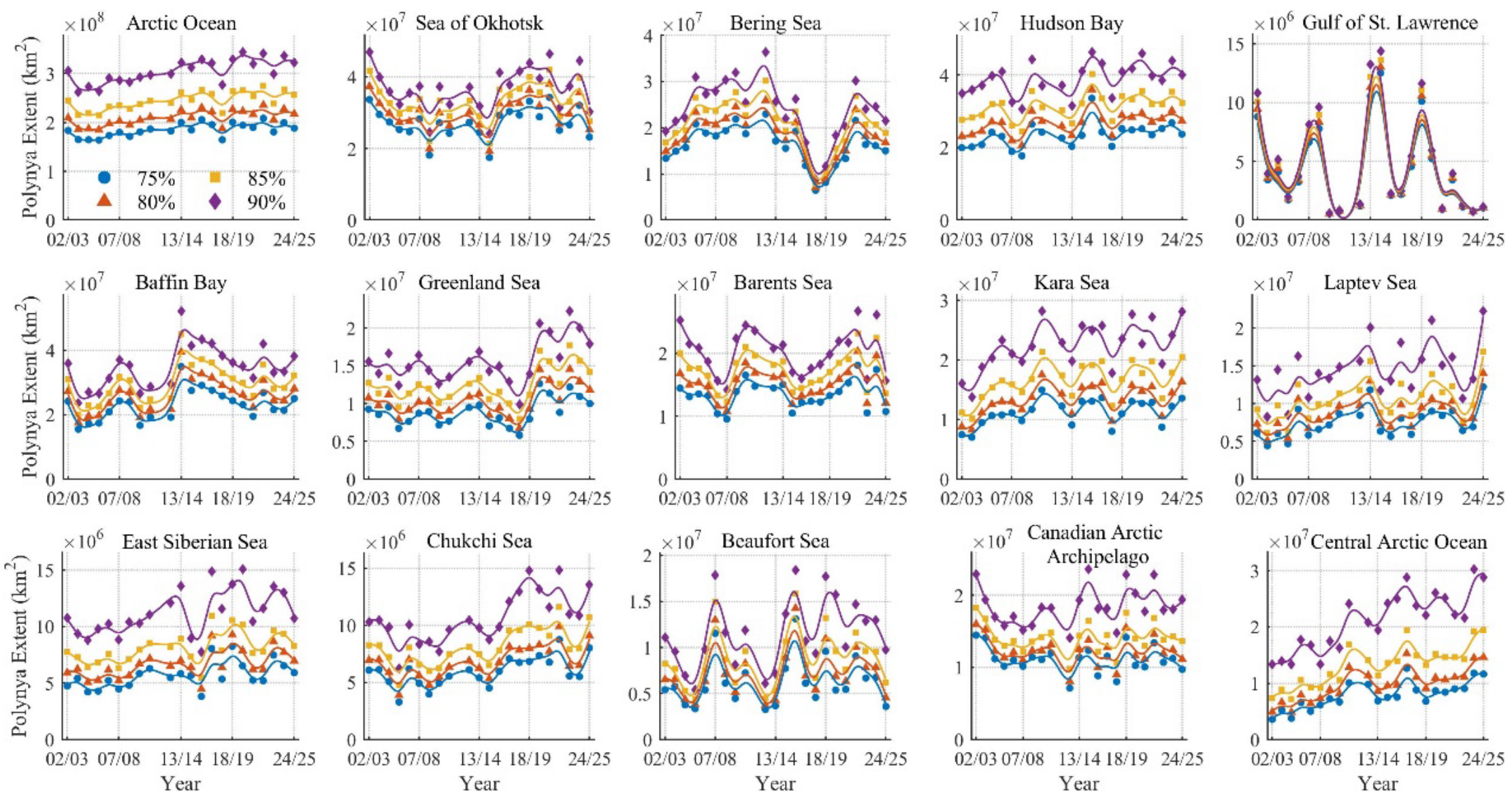

**Figure 4.** Yearly Arctic Polynya Extent Variation (2002/03-2024/25) based on different thresholds of SIC data (6.25 km). The missing 2011/12 season is not plotted.

Regional threshold sensitivity is highly uneven (Figure 5). The Central Arctic Ocean shows the largest relative response, while the Gulf of St. Lawrence shows very little response. The Kara, East Siberian, Greenland, Beaufort, and Laptev seas also show comparatively large responses, whereas the Okhotsk Sea, Canadian Arctic Archipelago, Bering Sea, Barents Sea, Baffin Bay, Hudson Bay, and Chukchi Sea are less sensitive in the reported comparison. Several regions show their strongest response in the 85%–90% interval. The sensitivity of SIC-defined extent is therefore geographically structured rather than a uniform Arctic-wide percentage uncertainty.

The regional pattern is consistent with the topology of a connectivity-based classifier. Raising TH not only admits additional boundary pixels; it can also join previously separated low-concentration patches or change their connection to the ocean-connected domain. Regions with many pixels near the upper threshold or with fragmented low-SIC geometry can therefore respond more strongly than regions whose mapped openings are sharply bounded. Morphological and geomorphon-based polynya studies similarly show that threshold and spatial context jointly control mapped boundaries[15], [30]. Figure 5 thus reveals a regional mapping sensitivity that arises from both the SIC distribution and the spatial organization of candidate pixels``111.

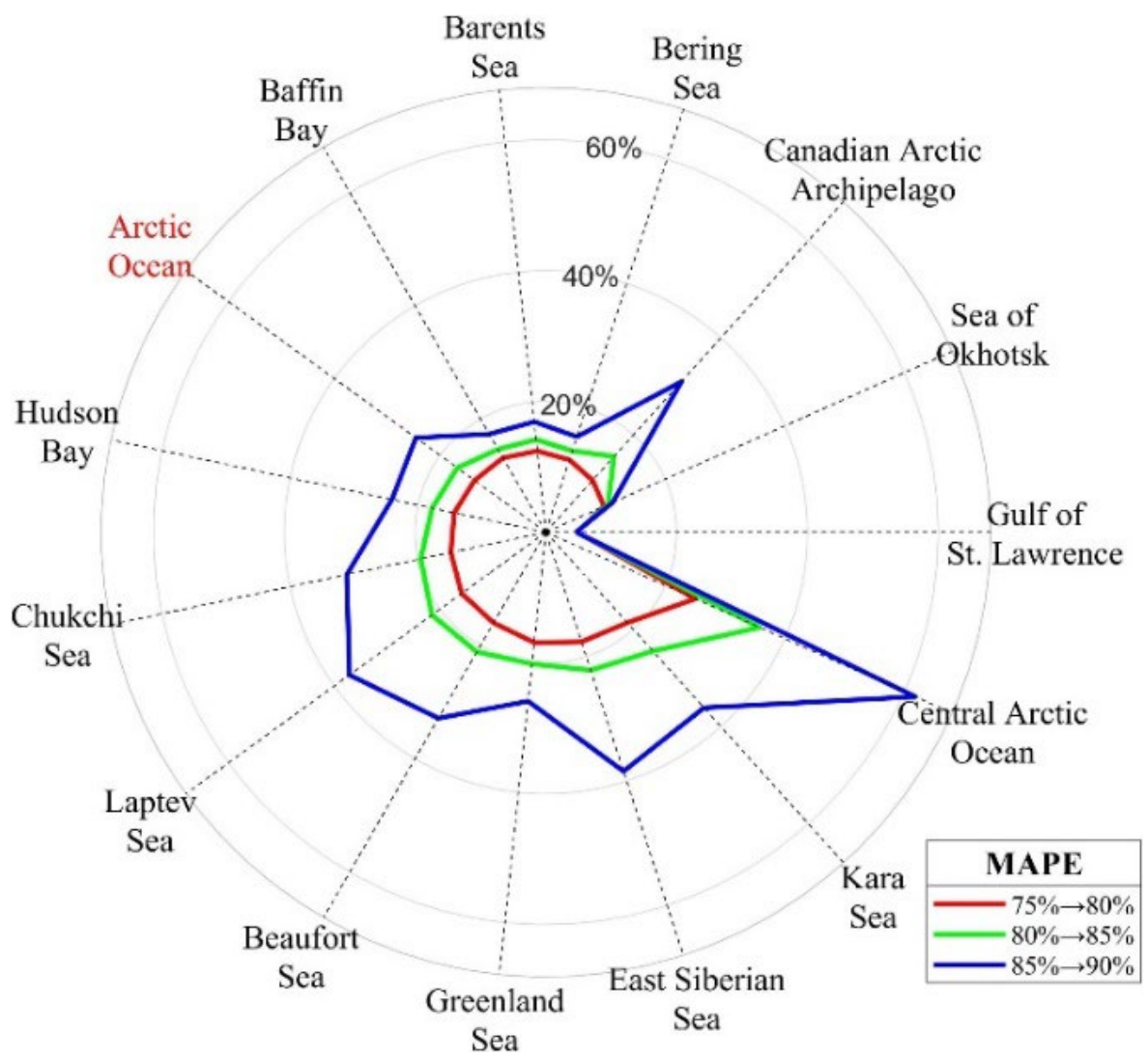

**Figure 5.** Regional relative differences between successive SIC-threshold settings: 75%–80%, 80%–85%, and 85%–90%. Each pair uses its lower threshold as the denominator.

Threshold sensitivity also has a clear seasonal structure (Figure 6). Aggregate separation among the four curves is comparatively small from September to December and becomes much more apparent from January onward. The threshold choice therefore affects late-winter and spring amplitude more strongly than early-season amplitude in the aggregate record. Comparable threshold-based studies have found that the spread among mapped polynya areas varies with the stage of polynya development and season[15], [16]. In the present record, this seasonal dependence reinforces the regional result from Figure 5: threshold uncertainty is not a single constant attached to the entire time series, but varies with both location and time of year.

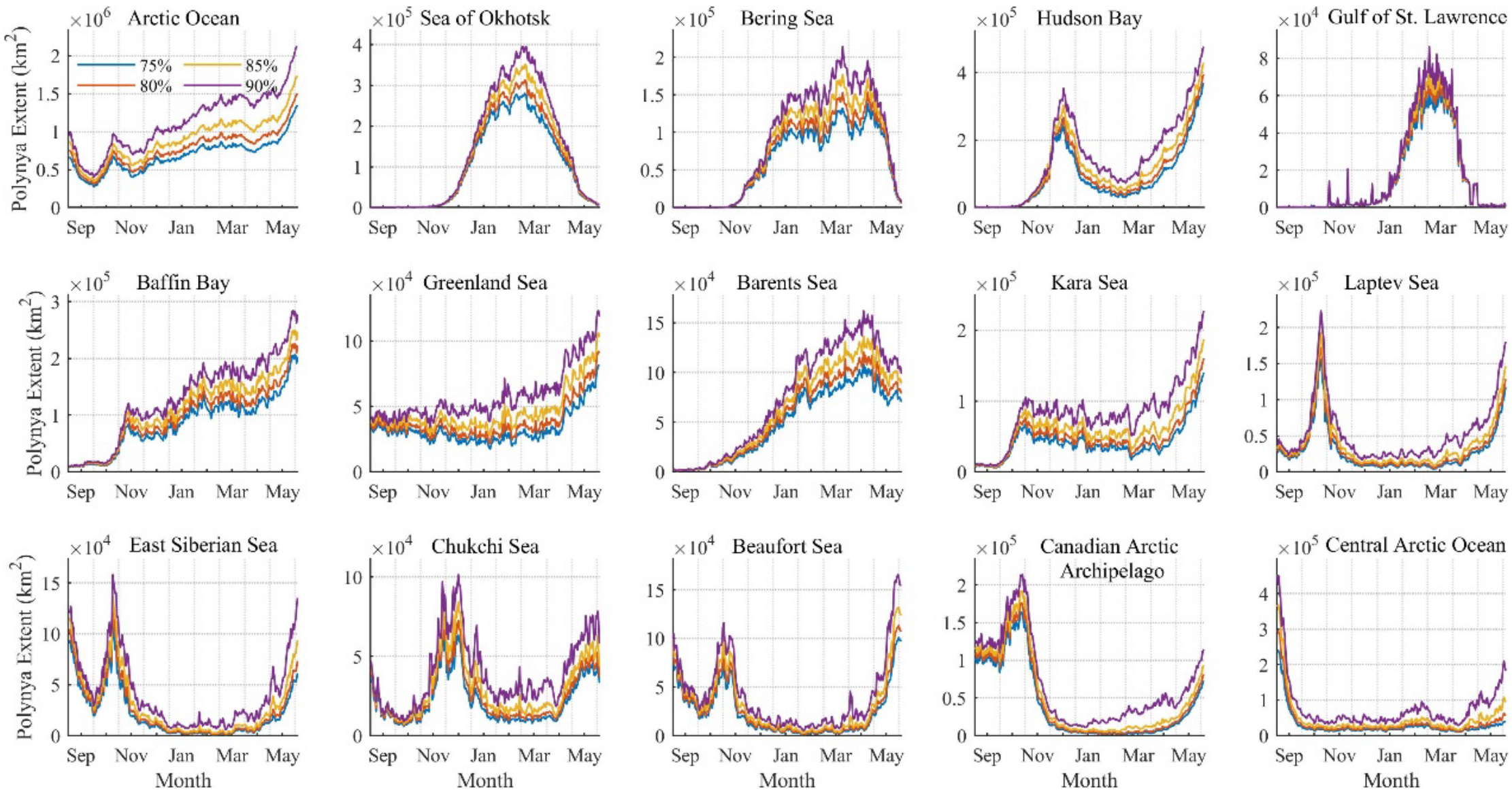

**Figure 6.** Monthly averaged polynya extent over multiple years derived using different sea ice concentration thresholds.

## 4.2 Sensitivity to SIC product selection

The product experiment fixes the upper SIC threshold at 80% and applies the same classification logic to the 6.25 km ASI, 12.5 km NT2, and 25 km NOAA/NSIDC inputs. Figure 7 compares their seasonal integrals over the available periods. Because the products differ in retrieval construction as well as grid spacing, this experiment measures practical product sensitivity rather than a controlled resolution-only effect. That distinction is retained while the results are interpreted in terms of the reproducibility of the mapped record.

Product choice changes the magnitude of the aggregate record substantially. The MAPE is 11.47% for the 12.5–25 km comparison over 2012/13–2024/25 and 68.19% for the 6.25–25 km comparison over the valid 2002/03–2024/25 pairs, excluding 2011/12. The 6.25 km series is generally larger in Figure 7. This direction is consistent with finer gridding retaining more small and fragmented openings, a behavior also reported in polynya sensitivity studies[5], [16], [19], while differences among the underlying SIC retrieval algorithms can independently alter concentration fields[6], [17], [18]. The large 6.25–25 km disagreement should therefore be interpreted as a combined product effect, not as grid spacing alone.

Regional product sensitivity is as heterogeneous as threshold sensitivity. Average product-comparison MAPEs are below 50% for Hudson Bay, the Kara Sea, and the Canadian Arctic Archipelago, but exceed 100% for the Greenland, Okhotsk, and Bering seas. This spread rules out an Arctic-wide rescaling that could convert one product's polynya extent into another's. It also agrees with broader SIC evaluations showing that algorithm differences vary across concentration regimes and surface conditions[6], [17], [18]. For coastal polynyas, where narrow or fragmented openings are common, spatial sampling and mixed pixels add another source of divergence[1], [19], [30]. The product experiment therefore identifies geography as a central dimension of uncertainty in mapped extent.

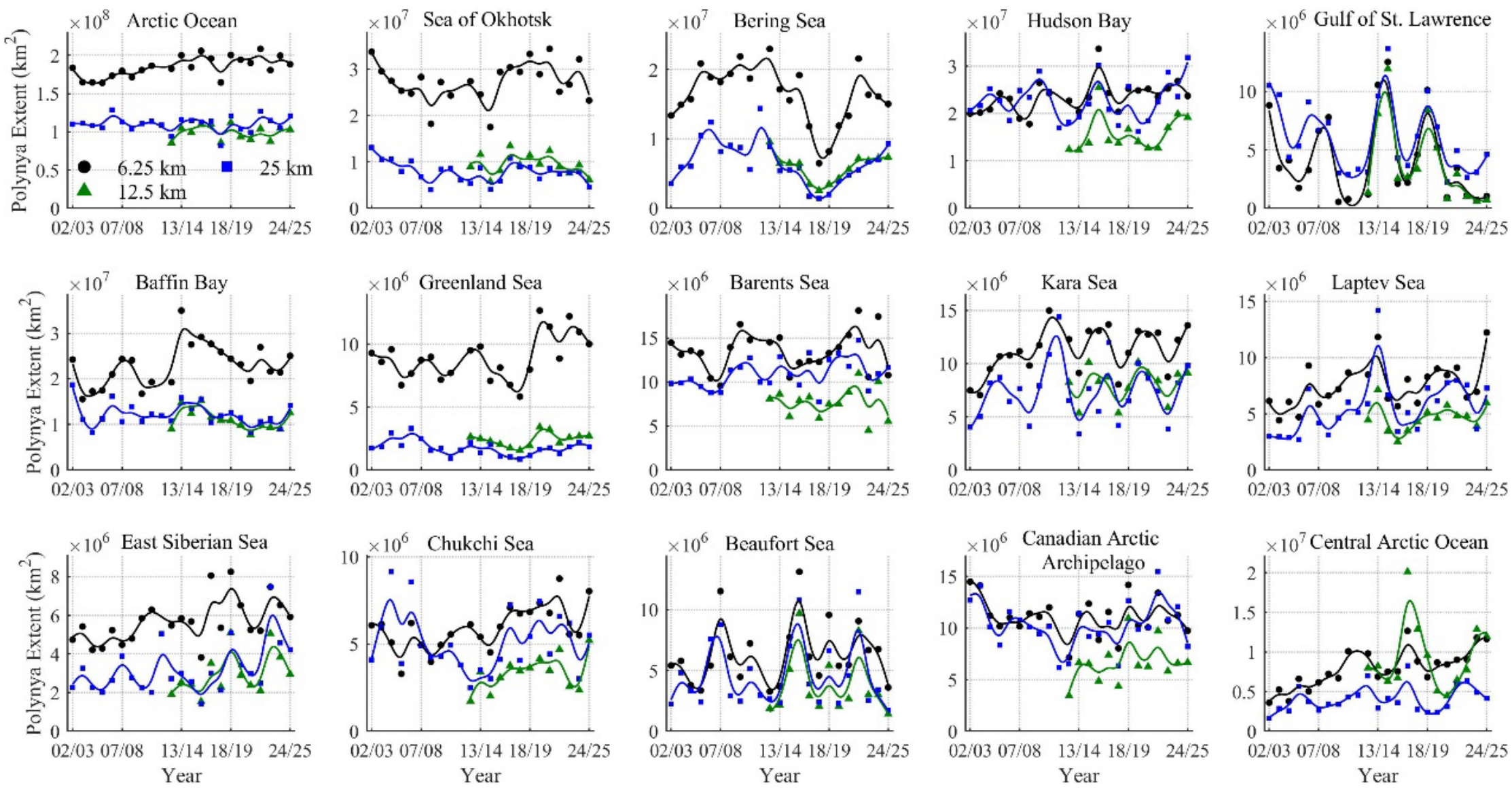

**Figure 7.** Interannual Arctic Polynya Extents (2002/03-2024/25) Derived from SIC Data at Different Resolutions.

Figure 8 adds a temporal dimension to the product comparison. Differences in the Okhotsk and Bering seas, Hudson Bay, and Baffin Bay are especially apparent from January through April. Yet several panels retain similar timing of major seasonal peaks even when their amplitudes differ strongly. Persistent amplitude offsets then accumulate into the seasonal integrals in Figure 7. As in the threshold experiment, the broad timing of activity can therefore be more stable than its absolute magnitude. This distinction helps explain why different SIC products may support a similar qualitative seasonal narrative while producing substantially different quantitative extent estimates.

Taken together, Figures 7 and 8 show that product selection affects both the magnitude and spatial distribution of mapped extent, with the strongest discrepancies concentrated in particular regions and parts of the season. Previous SIC intercomparisons document algorithm-dependent behavior at intermediate concentrations and under changing surface conditions[6], [17], [18], [31], [32], while polynya-specific comparisons demonstrate additional effects of grid spacing and morphology[15], [16], [19]. The present results fit this broader measurement picture: no single factor explains every difference, but the resulting configuration sensitivity is large enough that product identity must be treated as part of the definition of a long-term polynya record.

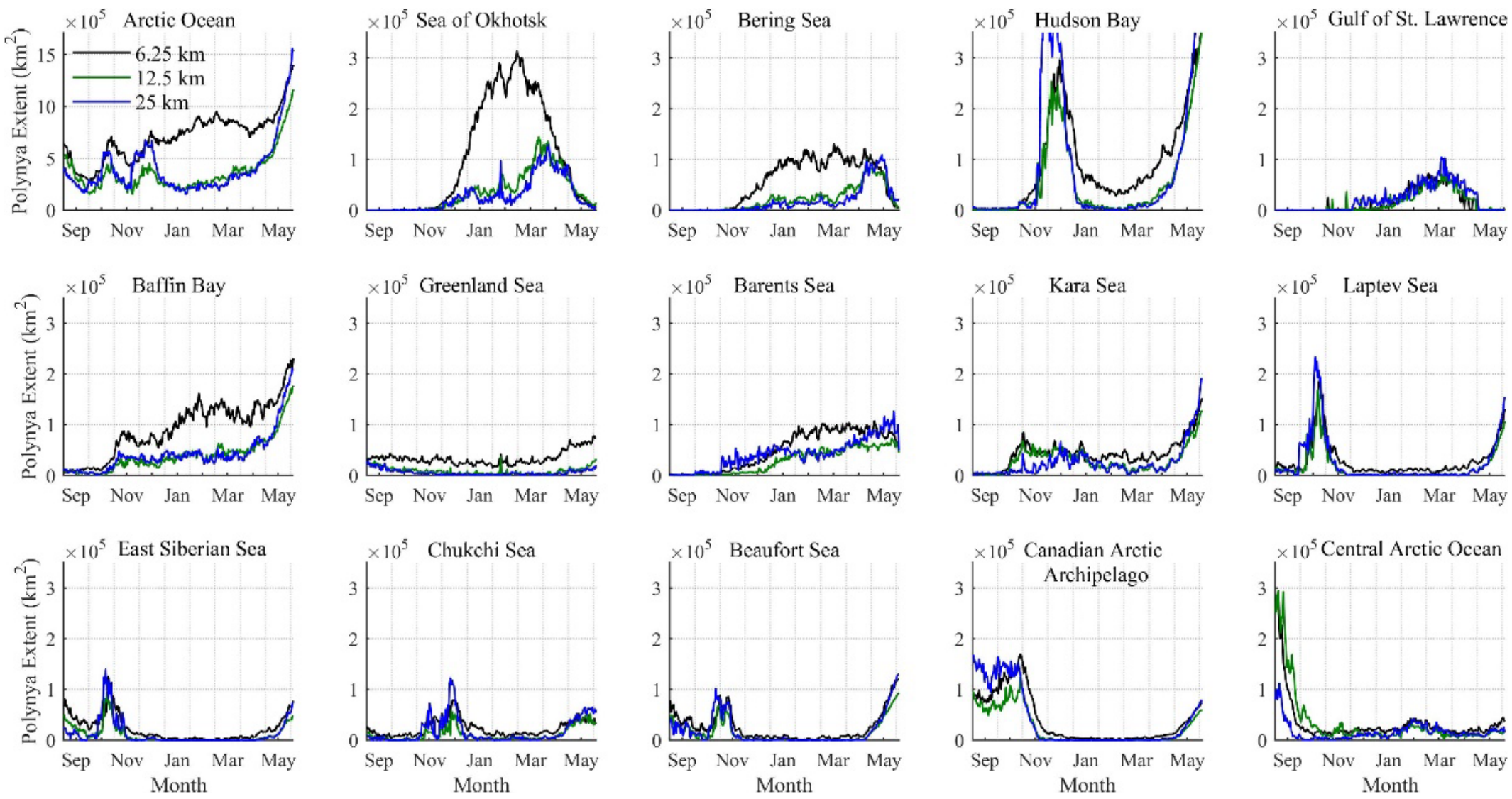

**Figure 8.** Multi-year Monthly Averaged Polynya Extents (2012/13-2024/25) Derived from SIC Data at Different Resolutions.

## 5 Discussion

### 5.1 Contribution of hierarchical partitioning and differentiated region growing

The methodological contribution of this study lies in combining hierarchical partitioning with differentiated region growing to resolve the spatial ambiguity between polynyas and the MIZ. Previous studies have used region-growing approaches to separate ocean-connected water from polynya-related features[1], while hybrid deep-learning methods have identified Arctic polynyas by subsequently removing ocean- and MIZ-related false detections[7]. The present framework separates the classification problem into two linked decisions: concentration defines the pixels eligible for classification, whereas connectivity and neighborhood context determine whether those pixels ultimately belong to a polynya or to the ocean-connected domain. This structure accommodates both detached openings within the consolidated ice pack and smaller candidate openings embedded near the broader ocean-connected low-concentration domain.

A key feature of the hierarchy is that the two SIC thresholds serve different purposes. The upper threshold defines the concentration range available to the classifier, whereas the fixed lower threshold helps identify open-water and candidate-polynya seeds within the ocean-connected low-SIC domain. Connected-component analysis provides the larger-scale spatial context, after which differentiated growth resolves the boundary between competing classes. Polynya growth accepts neighboring pixels whose SIC is no lower than that of the current growth pixel while maintaining separation from open-water seeds; subsequent open-water growth preserves pixels already assigned to polynyas. The resulting classification therefore combines local SIC transitions with larger-scale ocean connectivity. Importantly, zero-SIC pixels can remain classified as part of a polynya when they belong to an enclosed opening, a distinction that cannot be obtained from concentration values alone.

The examples in Figure 3 illustrate how these successive decisions generate spatially coherent polynya, MIZ, and open-water classes. Across the 63 assessed scenes, the mean overall accuracy of 96.74% and mean kappa of 0.93 indicate strong agreement with visual interpretation for the sampled polynya and MIZ classes. Applying the same workflow throughout the available September–May record provides a common operational basis for comparing SIC-defined openings across contrasting Arctic environments. More importantly, the explicit separation of candidate identification, boundary growth, and final classification makes the operational definition traceable, while the sensitivity experiments quantify how the resulting extent changes with threshold and SIC-product choice. Together, these elements provide a reproducible framework for interpreting regional contrasts and temporal variability in SIC-defined Arctic polynya extent.

### 5.2 Interpretation of threshold and product sensitivity

The two sensitivity experiments reveal a common structure. Their comparison conditions are summarized in Table 3. Absolute extent varies much more strongly across configurations than the timing of major temporal features, while the magnitude of this sensitivity differs substantially among regions and through the ice season. For threshold choice, aggregate differences grow from 13.48% to 65.04% across the tested settings while Figures 4 and 6 retain recognizable temporal organization. For product choice, MAPEs range from 11.47% to 68.19% under the stated comparisons, with regional values spanning from below 50% to above 100% in the reported groups. The combined message is therefore not simply that parameters matter, but that they affect the magnitude, geography, and seasonality of the mapped record in different ways.

**Table 3.** Comparison conditions for interpreting threshold and SIC-product sensitivity.

| Comparison | Quantity varied | Statistic (%) | Reference series | Valid seasons |
|---|---|---|---|---|
| 80% versus 75% | Upper SIC threshold | MAPE13.48 | 6.25 km ASI, TH = 75% | 2002/03–2024/25, excluding 2011/12 |
| 85% versus 75% | Upper SIC threshold | MAPE 32.47 | 6.25 km ASI, TH = 75% | 2002/03–2024/25, excluding 2011/12 |
| 90% versus 75% | Upper SIC threshold | MAPE 65.04 | 6.25 km ASI, TH = 75% | 2002/03–2024/25, excluding 2011/12 |
| 12.5 versus 25 km | SIC product; TH = 80% | MAPE 11.47 | 12.5 km NT2 | 2012/13–2024/25 |
| 6.25 versus 25 km | SIC product; TH = 80% | MAPE 68.19 | 25 km CDR | 2002/03–2024/25, excluding 2011/12 |

*Notes: All statistics compare seasonal extent integrals. MAPE follow Eq. (4), with the listed series as denominator. The threshold comparisons hold the input product fixed. The product comparisons change retrieval and sampling together. Their periods and denominators differ and are therefore interpreted within their stated comparison designs. Values summarize Sections 4.1–4.2.*

The threshold and product experiments should therefore be read as complementary rather than ranked against each other. Threshold experiments hold the observational input fixed and expose the sensitivity of the connectivity-based classifier to the admissible SIC range. Product experiments hold the classifier and 80% threshold fixed but change the SIC field itself. Together they identify two separate pathways by which a reported polynya extent can change: the operational boundary chosen within one SIC field and the observational product from which that field is derived.

The literature supports this two-pathway interpretation. SIC inter-comparisons show systematic differences among retrieval algorithms, especially away from compact-ice end members[6], [17], [18], while recent polynya studies demonstrate strong threshold, grid, and morphological sensitivity[15], [16], [19]. Kern et al.[19], for example, compared PSSM-derived polynya maps with higher-resolution AMSR-E 89-GHz ASI concentration maps and showed that differences in spatial resolution and observing frequency can lead to systematic differences in the retrieved open-water and thin-ice signatures, highlighting the sensitivity of polynya estimates to the characteristics of the underlying microwave product. The agreement in direction across independent studies strengthens the practical conclusion here: the processing configuration is part of the measurement definition, not a neutral post-processing detail.

For long-term Arctic monitoring, this means that consistency of configuration is essential for quantitative comparability. A fixed product version, upper SIC threshold, regional mask, geographic exclusions, and missing-data treatment should be reported with the extent record. The 80% threshold used here provides a reproducible operational reference within the tested range, while Figures 4–8 quantify how far results can move when that reference is changed. The more stable temporal features can support comparative analyses, but absolute extent values should remain tied to the configuration that produced them.

### 5.3 Interpretation of accuracy, sensitivity, and the physical meaning of SIC-defined polynya extent

The 63-scene assessment shows strong agreement between the mapped polynya and MIZ classes and visual interpretation of SIC imagery. Sampling therefore focused on the polynya and MIZ classes, whose assignment involves a genuine classification decision. By contrast, the high-concentration ice and open-water classes are generated deterministically by the workflow, so their classification carries no assignment error and they were not sampled separately. The reported accuracy therefore estimates the classification performance for the polynya and MIZ classes at the record scale under a stratified random sampling design. Independent higher-resolution observations, such as Landsat or SAR, would provide additional evidence on boundary placement and small features unresolved by the SIC grids[33]. The lower polynya producer's accuracies in December and April identify useful periods for targeted boundary assessment. Broader accuracy statements covering additional classes would require extending the sampling design accordingly[28], [29].

The sensitivity experiments quantify the effects of upper SIC threshold and product selection. Because the products differ in both retrieval algorithm and spatial sampling, their comparison captures a combined product effect. The 80% threshold provides a common reference for these experiments, although it has not been independently calibrated. Alternative neighborhood configurations and geographic exclusion rules were explored during method development, with the final settings selected based on the quality of the resulting polynya delineation and then held fixed throughout the sensitivity experiments. Their individual effects on mapping accuracy and extent have not been systematically quantified. Controlled regridding of a common SIC field, independent threshold calibration, and quantitative comparisons of alternative spatial rules would help distinguish these effects. Data continuity also warrants consideration: the primary record excludes the unavailable 2011/12 season, while interpolation across short gaps does not provide independent information on polynya opening or closure during those dates.

SIC-defined polynya extent represents the total area of pixels assigned to the polynya class, including both their open-water and ice-covered fractions. It therefore provides a consistent spatial basis for examining polynya distribution and temporal variability under a specified mapping configuration. Estimating heat flux or ice production requires additional information on ice thickness, open-water fraction, surface temperature, and atmospheric forcing[34], [35], [36]. Linking the mapped extent with these complementary observations would extend its use to quantitative assessments of polynya thermodynamic processes.

## 6 Conclusions

A pan-Arctic polynya mapping workflow has been developed using hierarchical

SIC partitioning, connected-component labeling, and differentiated growth of polynya and open-water regions. Applied to September–May seasons from 2002/03 to 2024/25, it separates enclosed low-concentration openings from the ocean-connected ice–water mixture and achieves a mean overall accuracy of 96.74% and a mean kappa coefficient of 0.93 in the 63-scene visual assessment.

The sensitivity experiments reveal the central result of the study: absolute SIC-defined polynya extent is highly configuration-sensitive, while broad temporal organization is comparatively more stable. Relative to 75%, aggregate threshold differences are 13.48%, 32.47%, and 65.04% at 80%, 85%, and 90%, respectively. The response is strongly regional and seasonal, with the largest threshold sensitivity in the Central Arctic Ocean, very weak sensitivity in the Gulf of St. Lawrence, and increasing aggregate separation from January onward. At a fixed 80% threshold, product-comparison MAPEs are 11.47% for 12.5 versus 25 km and 68.19% for 6.25 versus 25 km under their specified comparison periods and reference series. Regional disagreement ranges from below 50% in some masks to above 100% in others, showing that no universal Arctic-wide correction can reconcile configurations.

These findings shift the interpretation of sensitivity from a single error percentage to a structured property of the mapping system. Threshold and product choice alter magnitude in ways that depend on region and season, whereas many major temporal features remain recognizable. A long-term SIC-based polynya record is therefore most reproducible when product, threshold, masks, and connectivity rules are kept fixed and reported as part of the measurement definition. Within that framework, hierarchical partitioning and region growing provide an interpretable basis for comparing Arctic polynya extent through time.

**Funding:** This work was supported by the National Natural Science Foundation of China (No. 42376241)

**Acknowledgements**

We thank the University of Bremen and the National Snow and Ice Data Center

for providing the SIC products. We also acknowledge the Centre for Polar Observation and Modelling for distributing the regional mask.